\documentclass[10pt,twocolumn,letterpaper]{article}

\usepackage{wacv}              

\usepackage{wacv}
\usepackage{times}
\usepackage{epsfig}
\usepackage{graphicx}
\usepackage{amsmath}
\usepackage{amssymb}
\usepackage{booktabs}
\usepackage{algorithm2e}
\usepackage{algorithmic}

\usepackage[pagebackref,breaklinks,colorlinks]{hyperref}

\usepackage[capitalize]{cleveref}
\crefname{section}{Sec.}{Secs.}
\Crefname{section}{Section}{Sections}
\Crefname{table}{Table}{Tables}
\crefname{table}{Tab.}{Tabs.}

\def\wacvPaperID{1446} 
\def\confName{WACV}
\def\confYear{2024}

\begin{document}

\title{Semi-supervised Contrastive Outlier removal for Pseudo Expectation Maximization (SCOPE)}

\author{Sumeet Menon\\
University of Maryland, Baltimore County \\ University of Miami\\
1000 Hilltop Cir, Baltimore, MD 21250 \\
{\tt\small sumeet1@umbc.edu}
\and
David Chapman\\
University of Miami\\
1320 S. Dixie Highway, Coral Gables, FL. 33146\\
{\tt\small dchapman@cs.miami.edu}
}
\maketitle

\begin{abstract}
   Semi-supervised learning is the problem of training an accurate predictive model by combining a small labeled dataset with a presumably much larger unlabeled dataset.  Many methods for semi-supervised deep learning have been developed, including pseudolabeling, consistency regularization, and contrastive learning techniques.  Pseudolabeling methods however are highly susceptible to confounding, in which erroneous pseudolabels are assumed to be true labels in early iterations, thereby causing the model to reinforce its prior biases and fail to generalize to strong predictive performance.  We present a new approach to suppress confounding errors through a method we describe as Semi-supervised Contrastive Outlier removal for Pseudo Expectation Maximization
(SCOPE). Like basic pseudolabeling, SCOPE is related to Expectation Maximization (EM), a latent variable framework which can be extended toward understanding cluster-assumption deep semi-supervised algorithms.  
However, unlike basic pseudolabeling which fails to adequately take into account the probability of the unlabeled samples given the model, SCOPE introduces outlier suppression in order to identify high confidence samples for pseudolabeling.  Our results show that SCOPE improves semi-supervised classification accuracy for CIFAR-10 classification task using 40, 250 and 4000 labeled samples, as well as CIFAR-100 using 400, 2500, and 10000 labeled samples.  Moreover, we show that SCOPE reduces the prevalence of confounding errors during pseudolabeling iterations that would otherwise contaminate the labeled set in subsequent retraining iterations.
\end{abstract}

\section{Introduction}

Learning high quality model representations from limited labeled data is a problem that deep learning has not yet overcome. 
Although there have been much progress in this area, the vast majority of published deep learning algorithms need to be trained with large labeled data volumes in order to perform well.  
The applications for semi-supervised learning are numerous because in many domains, unlabeled data is plentiful yet high quality labeled data is scarce. Data labeling remains a task that is time consuming, expensive and error prone.  As such, methods to reduce the need for manually labeled data may be impactful. 

Semi-supervised learning becomes particularly challenging when the labeled data volumes are very small relative to the unlabeled volumes.  In this case, many methods, especially pseudolabel techniques are susceptible to confounding errors, in which the model at first learns some bias due to the small and inadequate labeled sample, and then proceeds to pseudolabel some of the data incorrectly with high confidence, and then proceeds to retrain treating incorrect pseudolabels as real labels thereby reinforcing its prior bias.  This confounding bias can prevent a pseudolabel learning algorithm from achieving acceptable performance.
\subsection{Intuition of the cause of confounding bias}
We now introduce our main intuition for what we believe to be a causal factor of confounding bias for pseudolabeling based semi-supervised Deep Neural-Network (DNN) models.  As part of the theoretical justification for the proposed SCOPE technique, we discuss the close relationship between pseudolabeling and Expectation Maximization (EM) ~\cite{dempster1977maximum}.
EM however is designed to be applied to generative models, whereas DNN classifiers making use of softmax final activation are within the category of discrimination models.  The use of a discrimination model with a EM-like meta-learning leads to mismatched assumptions, and more importantly, the potential for a confounding failure case in the presence of outliers.

\begin{figure}
	\centering
	\includegraphics[width=7cm]{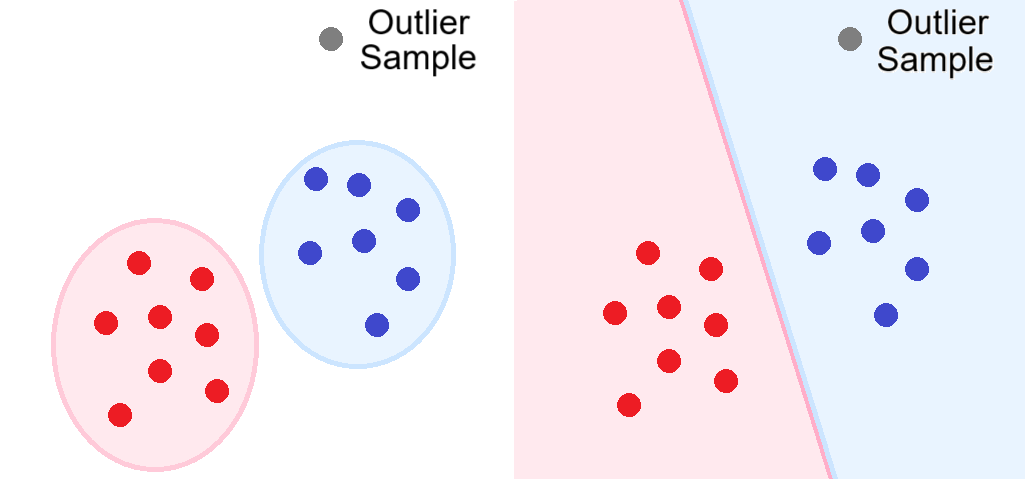}
	\caption{Illustration of incompatibility with cluster assumption. Left: generative model assigns low likelihood to outlier. Right: discrimination model confidently assigns outlier to blue cluster.
 }
 \label{fig1}
\end{figure}

Figure \ref{fig1} illustrates these mixmatched assumptions.
As seen in Figure \ref{fig1} (right), with a discrimination backbone, the outlier sample is incorrectly predicted to have high confidence of belonging to the blue cluster.  However, as seen in Figure \ref{fig1} (left), with a generative backbone, the outlier sample is correctly predicted to have low confidence of belonging to either cluster.
More precisely, the confident predictions are the ones with the greatest posterior probabilities 
$p(Y|X,\theta)$, whereas we describe as part of our theoretical justification that the most reliable samples
for EM have the
greatest posterior probability of the image/label tuple $p(Y,X |\theta)$.  As such, basic pseudolabeling ignores the probability of the unlabeled sample occurring given the model parameters $p(X|\theta)$. 
Clearly, if $X$ is an outlier sample then $p(X|\theta)$ is much smaller than if $X$ is an inlier sample.  As such, our intuition is that outlier samples can be understood as a potential failure case
that may exacerbate confounding.

\subsection{Contributions}

We summarize the main contributions of this paper as follows,

\begin{itemize}
  \item Combination of outlier removal and consistency regularization improves upon state-of-art results for semi-supervised classification.
  \item Removing outlier samples from semi-supervised pseudolabeling can reduce the number of confounding errors that contribute to confounding bias.
  \item Theoretical justification describing the relation between EM and Pseudolabeling which motivates incorporation of outlier removal strategies.
\end{itemize}


\section{Related Work}
The most basic pseudolabeling algorithm is self-training  ~\cite{yarowsky1995unsupervised, mcclosky2006reranking, olivier2006semi,zhai2019s4l,livieris2019predicting,rosenberg2005semi}.  With self-training, an initial classifier is first trained with a small amount of labeled samples and this initial classifier is used to label a subset of the unlabeled samples.
These pseudolabeled samples are then added to the training dataset for retraining if the pseudolabels pass a certain threshold.  Since the classifier was trained initially only with a small subset of labeled samples, self-training models are susceptible to the confounding bias problem.  Several papers have made use of an iterative form of self-training, in which the the model iteratively switches between pseudolabeling the unlabeled data, and retraining the model with confident pseudolabels added ~\cite{mustafa2020transformation, menon2020deep, nguyen2020active}. We call this iterative process \textit{basic pseudolabeling}, and this process can also be viewed as a specific interpretation of Expectation Maximization (EM) ~\cite{dempster1977maximum} with several heuristic assumptions added.  We review the close relationship between EM and basic pseudolabeling in detail as part of the theoretical justification section.

A related method to self-training is co-training which is a part of multi-view training where a dataset $S$ can be represented as 2 independent feature sets $S_1$ and $S_2$.
After the model is $m_1$ and $m_2$ are trained on the respective datasets, at every iteration, the predictions that surpass the predetermined threshold from exactly one model are then passed to the training dataset of the final model ~\cite{blum1998combining,prakash2014survey}.
In recent times, co-training has been used in 3-D medical imaging where the coronal, sagittal and axial view of the data was trained on three different networks ~\cite{xia20203d}.

A much more sophisticated pseudolabeling method is the Mean Teacher algorithm  ~\cite{tarvainen2017mean}. 
which takes exponential moving averages of the model parameters to obtain a much more stable target prediction which significantly improves convergence.
One of the drawbacks of this and related methods is the use of domain specific augmentations.
The need for domain-specific augmentation remains an open problem, although techniques such as Virtual Adversarial Training ~\cite{miyato2018virtual} can reduce the impact of this issue in certain domains.

Contrastive Loss is a prominent distance criteria for smoothness based semi-supervised deep learning techniques, and is the foundation for Momentum Contrast (MoCo), its successors and related approaches ~\cite{he2020momentum, oord2018representation, henaff2020data, hjelm2018learning, tian2020contrastive, misra2020self, li2020prototypical}.
There are several forms of contrastive loss ~\cite{hadsell2006dimensionality,wang2015unsupervised, wu2018unsupervised, hjelm2018learning} but in its most general form one must define a similarity loss $L_S$ to penalize similar samples from having different labels, as well as a difference loss $L_D$ to penalize different samples from exhibiting the same label.
Consistency Regularization is another smoothness based strategy that has led to MixMatch and its derivatives for image classification ~\cite{berthelot2019mixmatch,sohn2020fixmatch,berthelot2019remixmatch,mustafa2020transformation}.
Consistency regularization assumes that if one augments an unlabeled sample, it’s label should not change; thereby implicitly enforicing a
smoothness assumption between samples and simple augmentations thereof.
The most common distance measurement techniques for these purposes are Mean Squared Error (MSE), Kullback-Leiber (KL) divergence and Jensen-Shannon (JS) divergence
~\cite{jeong2019consistency,verma2022interpolation,yalniz2019billion,sajjadi2016regularization}.

\section{Basic Definitions}

We now provide a set of basic definitions that we refer to throughout this paper.  We provide additional clarification on the notation, including an annotated list of mathematical symbols, in supplementary materials.

\subsection{Problem Definition for Semi-Supervised Learning}
Semi Supervised learning can be defined as the problem of learning an accurate predictive model using a training dataset with very few labeled samples but a much larger number of unlabeled samples.
Let us say that we have a set of training samples $X$ and training labels $Y$, the samples can be further defined as a set of independent supervised samples $X_S$ and unsupervised samples $X_U$ along with supervised labels $Y_S$ and unobservable (latent) unsupervised labels 

In practice, the number of unsupervised unlabeled samples is also typically much larger than the number of supervised labeled samples $|X_S| \; <<  \; |X_U|$.

The accuracy of semi-supervised learning is typically evaluated using the multi-class classification task via cross validated benchmark on a withheld test set $X_T$ , $Y_T$.
It is further assumed that the training and test sets follow the same distribution of samples and labels. 
The goal of semi-supervised learning is to minimize the expected testing loss as follows.

\begin{equation}
min_\theta \ E(\; L(Y_T - \hat{Y}_T)\;; X_L\; , \; X_U, \;Y_L)
\end{equation}




\subsection{Definition of Basic Pseudolabeling}






We define \textit{basic pseudolabeling} as the process in which we add new images and pseudolabels of high confidence to the \textit{supervised} set, while simultaneously removing those images from the unsupervised set.  Define $\hat{Y}^t_H \subseteq \hat{Y}^t_U$ as the set of \textit{high-confidence} predictions, and $X^t_H \subseteq X^t_U$ as the associated high confidence unlabeled images, and $Y^t_H \subseteq Y^t_U$ as associated unobservable high confidence true labels at iteration $t$.

In the first iteration ($t=0$), the supervised pseudolabels are initialized to the respective true labels as follows,
\begin{equation}
\ddot{Y}^0_S = Y^0_S
\end{equation}
The expansion of the \textit{supervised} training set due to \textit{basic pseudolabeling} is given as follows,
\begin{align}
X^{t+1}_S &= X^t_S \cup X^t_H    \textit{\quad\quad\quad supervised images} \nonumber \\
Y^{t+1}_S &= Y^t_S \cup Y^t_H \textit{\quad\quad\quad supervised true labels} \\
\ddot{Y}^{t+1}_S &= \ddot{Y}^t_S \cup \hat{Y}^t_H \textit{\quad\quad\quad supervised pseudo-labels} \nonumber
\end{align}
In addition, the high confidence samples are removed from the unsupervised training set as follows,
\begin{align}
X^{t+1}_U &= X^t_U - X^t_H    \textit{\quad\quad\quad unsupervised images} \\
Y^{t+1}_U &= Y^t_U - Y^t_H \textit{\quad\quad\quad unsupervised true labels} \nonumber
\end{align}

\subsection{Definition of confounding errors and confounding bias}

It is important to note that although \textit{confounding bias} is recognized as an issue for semi-supervised bootstrapping methods such as pseudolabeling ~\cite{arazo2020pseudo}, we are unaware of a precise mathematical definition of \textit{confounding bias} and/or \textit{confounding errors}.  We propose a very simple definition of \textit{confounding errors} that is intuitive and easily measurable.

We define a \textit{confounding error} to be any sample with an erroneous pseudolabel that is included in the supervised set.  As such, sample $i$ is a \textit{confounding error} at timestep $t$ if and only if,
\begin{equation}
\ddot{Y}^t_{S,i} \neq Y^t_{S,i}
\end{equation}
Inclusion of an erroneous pseudolabel in the supervised set causes all henceforth iterations to mistakenly treat this sample as a true label during supervised learning.
We furthermore define \textit{confounding bias} as the difference between the observed model performance, and the hypothetical model performance that could be achieved if all \textit{confounding errors} were removed from the supervised set by an oracle in each iteration $t$ of pseudolabeling.


\section{Theoretical Justification}

\subsection{Relation between EM and Pseudolabeling}



Theoretically, pseudo-labeling as well as latent bootstrapping methods rely on the cluster assumption ~\cite{olivier2006semi}.
The cluster assumption can be intuitively paraphrased as follows: If there are 2 points that belong to the same cluster, then they (very likely) belong to the same class ~\cite{olivier2006semi}.
As such, clustering methods assume that the data is separable into $K$ clusters $C_1\ ...\ C_K$ in which the true decision boundary lies in-between the clusters, and does not pass through any individual cluster. 
%
Pseudolabeling is a special case of the cluster assumption, in which one not only assumes the decision boundary lies in-between the clusters, but further assumes the stronger condition that there is only one cluster per label category.
The EM algorithm ~\cite{dempster1977maximum} is the foundation for many clustering techniques.
Maximum likelihood estimation of simultaneous latent variable $\mathcal{Z}$ and model parameters $\theta$ can be obtained iteratively as follows
\begin{equation}
\begin{aligned}
Expectation:\quad\quad Q(\theta|\theta^t) &= E_{\mathcal{Z}|\mathcal{X}, \theta^t} \log L(\theta; \mathcal{X}, \mathcal{Z})\\
Maximization:\quad\quad\quad \theta^{t+1} &= argmax_\theta \; Q(\theta | \theta^t)
\end{aligned}
\end{equation}
We show that pseudolabeling is highly related to a specific interpretation of EM, in that the latent variable $\mathcal{Z}$ is defined as the unobservable (unsupervised) training labels $Y_U$.
Furthermore the observed variable $\mathcal{X}$ describes all observable data measurements available for training, including the supervised training data $X_S$, the supervised training labels $Y_S$, as well as the unsupervised training data as follows $X_U$,
\begin{equation}
\begin{aligned}
\mathcal{Z} = Y_U  \quad \quad  \mathcal{X} = X_S , Y_S , X_U
\end{aligned}
\end{equation}
Given the basic statistical identity that $L(a|b)=p(b|a)$, the Expectation step under this interpretation is presented as follows.
\begin{equation}
Q(\theta|\theta^t) = E_{Y_U|X_S,Y_S,X_U,\theta^t} \log p(X_S, Y_S, X_U, Y_U|\theta)
\end{equation}

One must further assume sample independence of the individual samples lying within the training dataset.
Under this common assumption, the supervised and unsupervised contributions to the maximum likelihood expectation step can be split additively and simplified as follows,
%
%
%
\begin{equation}
\begin{aligned}
Q(\theta|\theta^t) =& \\
\texttt{\small{superivsed branch}}\quad\quad & \log p(X_S, Y_S|\theta) \\
\texttt{\small{unsuperivsed branch}}\quad\quad &+ E_{Y_U|X_U,\theta^t} \log p(X_U, Y_U|\theta) 
\end{aligned}
\end{equation}
One can also apply an additional Bayesian identity in that $p(a,b | c) = p(a|b,c) p(b| c)$.
As such, the expectation can be expanded as follows,
\begin{equation}
\begin{aligned}
Q(\theta|\theta^t) &= \log p(Y_S|X_S, \theta) \ p(X_S|\theta) \\
&+ E_{Y_U|X_U,\theta^t} \log p(Y_U| X_U, \theta) \ p(X_U| \theta)
\end{aligned}
\label{eq13}
\end{equation}
The expected value term in equation \ref{eq13} is difficult to evaluate.  As such, a reasonable approximation is to define the unlabeled predictions $\hat{Y}_U=E_{Y_U|X_U,\theta^t}(Y_{U})$ and to attempt to distribute this term within the expected value.
One may furthermore expand this formula into summation terms over the supervised samples $N$, unsupervised samples $M$, and classes $C$ as follows.  In this expanded form, $Y_S$ and $\hat{Y}_U$ are written respectively as one hot-encoded, and predictive probability tensors of dimensions $[N,  C]$ and $[M, C]$.  The use of one-hot notation necessitates multiplication by term $Y_{S,i,c}$ which is non-zero only for the target class.  Analogous multiplication is required for the unsupervised term $\hat{Y}_{U,i,c}$.  We also define weighting terms $w_{S,i}$, and $w_{U,i}$ as simple variable substitutions of $p(X_{S,i}|\theta)$ and $p(X_{U,i}|\theta)$.  This approximation of equation \ref{eq13} expands as follows,

\begin{equation}
\begin{aligned}
    Q(\theta|\theta^t) &\approx \\
	& \quad \texttt{\footnotesize{superivsed weighted log loss}}\\
	& \sum_{i=1}^{N} \sum_{c=1}^{C} w_{S,i} Y_{S,i,c} \log p(Y_{S,i,c} | X_{S,i}, \theta)\\
	& \quad \texttt{\footnotesize{unsupervised weighted log loss}}\\
	+& \sum_{i=1}^{M} \sum_{c=1}^{C} w_{U,i} \widehat{Y}_{U,i,c} \log p(\widehat{Y}_{U,i,c}|X_{U,i},\theta) \\
 \text{where}
    \\
    &w_{S,i} = p(X_{S,i}|\theta) \text{ and}\ w_{U,i} = p(X_{U,i}|\theta)
\end{aligned}
\label{eq15}
\end{equation}
When written in this form, the similarity between EM and pseudolabeling becomes apparent in the sense that a weighted version of well-known log-loss appears not only for the labeled samples, but also for the unsupervised pseudolabeled samples.

There are notable differences however.  One apparent difference is that for the EM approximation of equation \ref{eq15}, $\hat{Y}_{U,i}$ refers to a predicted probabilities, whereas in pseudolabeling, this term is fully binarized via $argmax$ over the $C$ label categories.
The biggest difference between pseudolabeling and EM arises however, when looking at the terms $w_{S,i}$ and $w_{U,i}$ in equation \ref{eq15}.  Notice that in a true EM framework, these terms would need to take the form $p(X|\theta)$, but evaluating these terms accurately may be impractical for \textit{discrimination models}, which attempt to directly calculate predicted probabilities of the form $p(Y|X,\theta)$, skipping the term $p(X|\theta)$.  We believe this mathematical dilemma is related to the intuitive dilemma that we highlight in Figure 1, in that that by ignoring $p(X|\theta)$, one has no sense of whether any given sample is an outlier.

It is also notable that the absolute magnitude of the $w_{S,i}$ and $w_{U,i}$ terms has no impact on the gradient direction, only the relative weighting of these terms affects gradient direction.  So if evaluating these sample weights directly is impractical, how might a discrimination pseudolabeler go about assigning weights to train a semi-supervised model in practice?  Standard practice is to employ a heuristic weighting scheme which may be empirically rooted.  
As such, it can clearly observed that \textit{basic pseudolabeling} is highly related to EM, but due largely to this heuristic weighting cannot be considered a true \textit{EM} approximator.

The simplest practical heuristic would be to omit the weights all-together by assigning $w_{S,i} = w_{U,i} = 1$.  Doing so would result in a strategy in which the pseudolabels of all unlabeled samples are treated as true labels in every single iteration, thereby yielding a simple log-loss across all fully-supervised, and pseudolabeled samples.  Such a simplistic approach is easy to implement, but likely to cause a lot of confounding, because \textit{low-confidence} pseudolabeles will be treated as true labels in every single iteration.

\textit{Basic pseudolabeling} has a better \textit{meta-heuristic}, by iteratively adding \textit{high confidence} samples to the supervised set in iteration $t$.  In the framework of equation \ref{eq15}, this iteration would be the equivalent of starting with all \textit{supervised} weights $w_{S,i} = 1$, and all \textit{unsupervised} weights $w_{U,i} = 0$.  After every iteration, a set of new \textit{high confidence} unlabeled samples is selected, and their weights are then flipped from $0$ to $1$.  We rewrite \ref{eq15} with this "growing" set of \textit{supervised} samples, using the \textit{basic pseudolabeling} notation from section 3.2 as follows,
\begin{equation}
\begin{aligned}
    Q(\theta|\theta^t) \approx & \sum_{i=1}^{N} \sum_{c=1}^{C} \ddot{Y}^t_{S,i,c} \log p(\ddot{Y}^t_{S,i,c} | X_{S,i}, \theta)\\
\text{where}& \\
    \ddot{Y}^{t+1}_{S} = & \ \ddot{Y}^t_S \cup \hat{Y}^t_H \quad \text{and} \quad \hat{Y}^t_H \subseteq  \ \hat{Y}^t_U
\end{aligned}
\label{ref16}
\end{equation}
But how does one select \textit{high confidence} pseudo-labels?  The most common approach for 
\textit{basic pseudolabeling} is to select samples with high predictive probabilities where $p(Y|X,\theta^t)$ is large.  But this common approach ignores outliers.  Instead, the SCOPE strategy selects the inlier samples, where $p(X|\theta^t)$ is large.  This inlier-selection (thus outlier removal) approach has greater resemblance to strict EM, which defines $w_{U,i}=p(X_{U,i}|\theta)$ to be the optimal criteria.

\section{Methodology}

\begin{figure*}[h]
	\centering
	\includegraphics[width=0.95\textwidth]{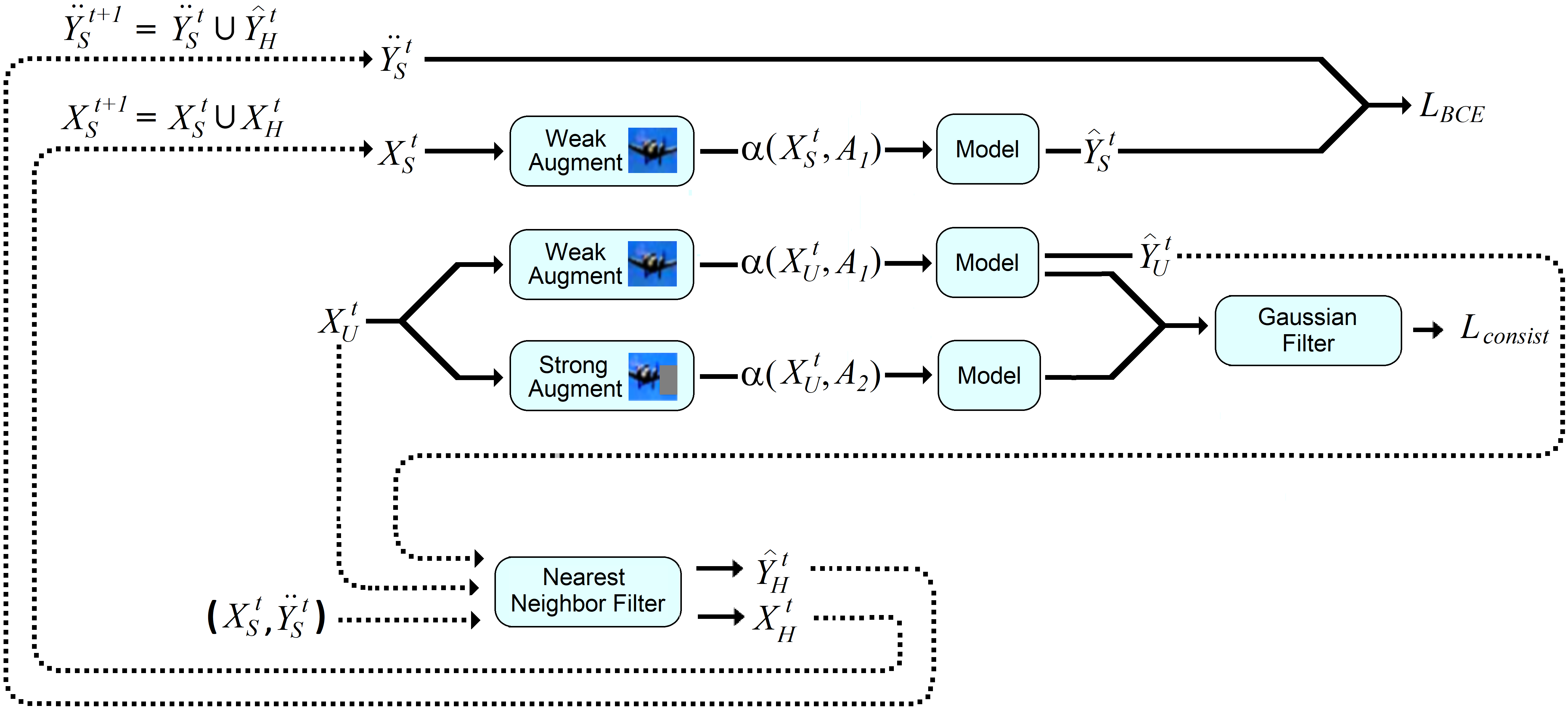}
	\caption{Description of the SCOPE Meta-learning algorithm.}
	\label{fig2}
\end{figure*}


Figure \ref{fig2} shows the SCOPE meta-learning algorithm.  The solid black lines show the portion of the SCOPE algorithm that is included within a single gradient update.  The dotted black lines show the portion of the algorithm that is an external pseudolabeling procedure that is calculated once per epoch in order to expand the supervised image $X^t_S$ and pseudolabel $\ddot{Y}^t_S$ sets.
The top portion of the figure shows the supervised path, exhibiting the Binary Cross Entropy loss term $L_{BCE}$.  This path also includes weak augmentation to help with convergence given the low sample size of the supervised set especially in early iterations.
The middle portion of this figure shows the unsupervised path, which highlights the calculation of the consistency regularization loss term $L_{consist}$ which compares the result of weak and strong augmentation of the unsupervised images.  Notice, that the Gaussian filter for outlier suppression is also included within the unsupervised path toward the weighting of the $L_{consist}$ term.
The bottom portion shows the pseudolabel updating portion of the SCOPE algorithm.  This portion includes the Nearest Neighbor filter which is used to select high confidence images $X^t_H$ and pseudolabels $\hat{Y}^t_H$.  These high confidence images and pseudolabels are used to expand the supervised set for the next iteration of pseudolabeling.

\subsection{Consistency Regularization}

Consistency regularization is an approach in which one adds an additional constraint that the predicted label should not change through augmentation.  Let us define A as the space of augmentation parameters, and $\alpha$  as the augmentation function.  Consistency regularization introduces the following constraint.

\begin{equation}
\begin{aligned}
p(Y|\alpha(X,A_1),\theta) = p(Y|\alpha(X,A_2),\theta)    
\\
\text{  for all   $A_1$ , $A_2$  $\in$  A}
\end{aligned}
\end{equation}

Consistency regularization is often defined with  as a random function, but it is equivalently presented here with  as a deterministic function but randomly chosen parameters $A_1, A_2 \in  A$  , where $A$ is the space of augmentation parameters. This constraint states algebraically that augmentation should not change the predicted labels.  Consistency regularization, like other forms of regularization, can be implemented by adding a penalty $L_{consist}$ to the overall loss function for optimization as follows,  

\begin{equation}
\label{eq18}
\begin{aligned}
L_{consist}=&E_{A_1 , A_2  \in  A} \\ & \quad L\big( p\left(Y|\alpha(X, A_1), \theta\right) \\& \quad \quad - p\left(Y|\alpha(X, A_2), \theta\right) \big)
\end{aligned}
\end{equation}

Consistency regularization is highly dependent on the ability to obtain a viable augmentation function that is unlikely to alter the true label of the image.  Recent work in the use of Consistency regularization for semi-supervised learning has yielded a number of augmentation functions that perform well for image and digit classification datasets as described by the following papers: ~\cite{sohn2020fixmatch,berthelot2019mixmatch,berthelot2019remixmatch}.  SCOPE makes use of Control-Theory Augment (CT-Augment) ~\cite{berthelot2019remixmatch}, Cut-out Augment ~\cite{CutAugment}, and Rand Augment ~\cite{RandAugment}.




\subsection{Gaussian Filtering Outlier Suppression}

Gaussian filtering is used for outlier suppression of the the consistency regularization term $L_{consist}$ as seen in Figure \ref{fig2}.  Both the Gaussian filter as well as the Nearest Neighbor filter are designed to identify unlabeled samples that reside within a \textit{high density} space given the present model parameters such that $p(X^t_U,i|\theta^t) > \tau$.  The primary advantage of the Gaussian filtering method is that it is relatively straightforward to integrate within the gradient descent portion of the training loop (Fig \ref{fig2} black lines), as it does not require any double loops as is the case for the Nearest Neighbor filter. For this reason the Gaussian filter is applied to suppress outliers within the unsupervised $L_{consist}$ term, whereas the Nearest Neighbor filter is applied at the end of each epoch to select \textit{high confidence} samples $X^t_H$ and pseudolabels $\hat{Y}^t_H$ to expand the supervised set. 

The Gaussian filter identifies the \textit{high density} regions where $p(X|\theta^t) > \tau$ by 
assuming a parametric distribution to directly measure
$p(X|\theta^t)$ and comparing it to a threshold $\tau$.  This parametric distribution assumes one Gaussian cluster per label category.

For simplicity the Gaussian filtering technique for SCOPE uses the output predicted probabilities as the manifold $F(X, \theta^t)$, and furthermore assumes use of the diagonal joint multivariate normal distribution as the distribution of output features $F(X, \theta^t)$.
We initialize the mean $\mu_c^t$ and the standard deviation $\sigma_c^t$ for the probability distribution $F(X, \theta^t)$ where $Y=c$.
\begin{equation}
\begin{aligned}
p(X|\theta^t) = \frac{1}{C} \sum_{c=1}^{C}  \frac{1}{\sqrt{2 \pi \sigma^t_c}} \exp\left(\frac{(F(X, \theta^t) - \mu_c^t)^2} {(\sigma_c^t)^2}\right)
\end{aligned}
\end{equation}
Theoretically, the $p(X|\theta^t)$ requires summing across all clusters, but in practice, it is adequate to calculate $p(X|\theta^t)$ only for the assigned cluster $c_i = \hat{Y}^t_{U,i}$.  This is because, in practice, if a sample is an inlier of the overall probability density, then it should be an inlier of its primary assigned cluster as well.

The pseudocode for the Gaussian filter is shown in Algorithm \ref{alg1}.  The output of this algorithm is a boolean vector $g_i$.  The unsupervised samples for which $g_i=1$ are considered inliers for which consistency regularization is applied, and the unsupervised samples for which $g_i=0$ are considered outliers, for which the consistency regularization is pruned and does not affect training gradients.
\begin{algorithm}
\DontPrintSemicolon
\caption{Gaussian Filter}
    \For{$c$ in $1...C$} {
        $\mu^t_c = E\left(F(X^t_U,\theta)|\hat{Y}^t_U=c\right)$ \\
        $\sigma^t_c = E\left(\left(F(X^t_U,\theta)-\mu^t_c\right)^2|\hat{Y}^t_U=c\right)$
    }
    \For{$i$ in $1...N_U$} {
        $c=\hat{Y}^t_{U,i}$\\
        $p^t_{U,i} = \frac{1}{\sqrt{2 \pi \sigma^t_c}} \exp\left(\frac{(F(X, \theta^t) - \mu_c^t)^2} {(\sigma_c^t)^2}\right)$ \\
        \eIf{$p^t_{U,i} > \tau$} {
            $g_i = 1$
        } {
            $g_i = 0$
        }
    }

\label{alg1}
\end{algorithm}
\begin{table*}
	\caption{CIFAR-10 Accuracy Results}
	\label{acc-table1}
	\centering
	\begin{tabular}{llll}
		\toprule
		Method & 40 labels  & 250 labels  & 4000 labels  \\
		\midrule
		$\pi$-Model \cite{rasmus2015semi} &  & $45.74\% \; (44.76,46.72)$ & $85.99\% \; (85.29,86.66)$  \\
		Pseudo-label \cite{leepseudo} & & $50.22\% \; (49.24,51.20)$ & $83.91\% \; (83.17,83.63) $  \\
		Mean Teacher \cite{tarvainen2017mean} & & $67.68\% \; (66.75,68.60)$ & $90.81\% \; (90.23,91.37)$  \\
		Mix-Match \cite{berthelot2019mixmatch} & $52.46\% \; (51.48, 53.44)$ & $88.95\% \;  (88.32,89.56)$ & $93.58\% \; (93.08,94.05)$  \\
		Fix-Match \cite{sohn2020fixmatch} & $88.61\% \; (87.97, 89.23)$ & $94.93\% \; (94.48,95.35)$ & $95.69\% \; (95.27,96.08)$  \\
		\textbf{SCOPE} & $\boldsymbol{93.58\%  (93.08, 94.05)}$ & $\boldsymbol{95.52\%  (95.10,95.92)}$  & $\boldsymbol{95.82\% (95.41,96.20)}$  \\
		\bottomrule
	\end{tabular}
\end{table*}
\begin{table*}
	\caption{CIFAR-100 Accuracy Results}
	\label{acc-table2}
	\centering
	\begin{tabular}{llll}
		\toprule
		Method & 400 labels  & 2500 labels & 10000 labels \\
		\midrule
		$\pi$-Model  \cite{rasmus2015semi} &  & $42.75\% \; (41.78, 43.73)$ & $62.12\% \; (61.16, 63.07)$   \\
		Pseudo-label \cite{leepseudo} & & $42.62\% \; (41.65, 43.60)$ & 63.79\% (62.84, 64.73)   \\
		Mean Teacher \cite{tarvainen2017mean} & & $46.09\% \; (45.11, 47.07)$ & 64.17\% (63.22,65.11)  \\
		Mix-Match \cite{berthelot2019mixmatch} & $32.39\% \; (31.47, 33.32)$ & $60.06\% \;  (59.09, 61.02)$ & 71.69\% (70.80, 72.57)  \\
		Fix-Match \cite{sohn2020fixmatch} & $50.55\% \; (49.57, 51.53)$ & $71.36\% \; (70.46, 72.24)$ & 77.4\% (76.57, 78.22)  \\
		\textbf{SCOPE} & $\boldsymbol{57.81\% \; (56.83, 58.78)}$ & $\boldsymbol{72.98\% \; (72.10, 73.85)}$  & $\boldsymbol{78.38\% \; (77.56, 79.18)}$ \\
		\bottomrule
	\end{tabular}
\end{table*}
\subsection{Nearest Neighbor Outlier Removal}
Nearest Neighbor filtering is employed for the selection of \textit{high confidence} samples during the pseudolabel updating branch of the SCOPE architecture as seen in Figure \ref{fig2}.  Intuitively, if a sample $X$ is nearby other supervised samples in $X^t_S$, then one might determine that $p(X|\theta^t) > \tau$.  Conversely, if a sample $X$ is far from all known supervised samples in $X_S$ one may conclude that $p(X |\theta^t)< \tau$.  
This intuition has a theoretical basis, because the K-Nearest Neighbor algorithm is the optimal classifier assuming a Variable-width Balloon Kernel Density Estimator. 
%
%
It can be shown that under these assumptions, $p(X | \theta ^ t)$ increases monotonically as sum of euclidean distances to the k nearest samples $d(X, X_s, k)$ decreases as follows,
\begin{equation}
\label{eq24}
\begin{aligned}
\forall \; \tau \;\;\;\;\;\;\;  \exists \; \gamma \;\;\; \text{s.t.}
\\
 p(X | \theta^t)> \tau   \;\;\;   iff  \;\;\;     d(X, X_s, k) \; < \;\; \gamma
\end{aligned}
\end{equation}
The Nearest Neighbor outlier removal step for the SCOPE algorithm is based on this strategy with three adjustments.  First, rather than euclidean distance between sample features, cosine similarity is used.  This is because cosine similarity is a standard similarity metric for latent sample-to-sample comparisons, and is particularly common for contrastive learning techniques.  Secondly, rather than comparing distance to all samples within the supervised set, this comparison is made only for the samples within the predicted pseudo-label category.  Thirdly, in accordance with equation \ref{eq24}, we need to make use a maximum cutoff distance $\gamma$ in order for the Nearest Neighbor filter to identify a \textit{high density} space.  Because \textit{cosine similarity} is a similarity metric, rather than a distance, the maximum distance $\gamma$ represents a minimum cosine similarity.  Algorithm \ref{alg2} shows the pseudocode for the Nearest Neighbor filtering algorithm to select \textit{high confidence} samples.

\begin{algorithm}
\DontPrintSemicolon
\caption{Nearest Neighbor Filter}
    $X^t_H = \emptyset \; \text{,} \ddot{Y}^t_H = \emptyset$ \\
    \For{$i$ in $1...N^t_U$} {
        $f_{U,i} = F(X^t_{U,i},\theta^t)$ \\
        $class = \hat{Y}^t_{U,i}$ \\
        $n_{similar} = 0$ \\
        \For{$j$ in $1...N^t_S$} {
            \If{$\ddot{Y}^t_{S,j}==class$} {
                $f_{S,i} = F(X^t_{S,i},\theta^t)$ \\
                $cos\_sim = \frac{f_{S,i} \cdot f_{U,i}}{||f_{S,i}|| \; ||f_{U,i}||} $ \\
                \If {$cos\_sim > \gamma$} {
                    $n_{similar} = n_{similar} + 1$
                }
            }
        }
        \If {$n_{similar} \ge k$} {
            $X^t_H = X^t_H \cup \{ X^t_{U,i}\}$ \\
            $\ddot{Y}^t_H = \ddot{Y}^t_H \cup \{ \hat{Y}^t_{U,i}\}$
        }
    }
\label{alg2}
\end{algorithm}

%
%
%
%
%
%


\section{Experimental Setup, Results and Ablation Study}

The SCOPE test accuracy was evaluated on the CIFAR-10 and CIFAR-100 datasets.  
This accuracy was compared against the following state-of-the-art semi-supervised learning algorithms, $\pi$-Model \cite{rasmus2015semi}, Pseudo-label \cite{leepseudo}, Mean-Teacher \cite{tarvainen2017mean} Mix-Match \cite{berthelot2019mixmatch} and Fix-Match \cite{sohn2020fixmatch}. 
All of the models in this comparison make use of the Wide ResNet architecture for the classifier. The SCOPE model is initially trained for 1024 epochs without the pseudolabeling procedure which expands the \textit{supervised} set with \textit{high confidence} samples.  After 1024 iterations, the model begins adding pseudolabels, and the \textit{supervised} set begins to grow.
All of the hyper-parameters information is available in the supplementary materials.
This paper evaluates the testing outcomes across three specific configurations of labeled and unlabeled data for each of these datasets.

Both the CIFAR-10 and CIFAR-100 datasets exhibit 50000 training samples and 10000 testing samples.
For CIFAR-10, three configurations were evaluated consisting of 40, 250, and 4000 labeled samples. Similarly, for CIFAR-100, the three experiments involve 400, 2500, and 10000 labeled samples. For experiments on both of these datasets with different configuration of labeled samples, the remaining samples in the training dataset were used as unlabeled samples.
In the following experiments, we have used the Clopper-Pearson confidence interval technique specifically to assess the classification accuracy.

For CIFAR-10 with 40 labeled samples, SCOPE significantly and substantially outperforms Mix-match and Fix-Match, and achieves accuracy of 93.58\% (CI: 92.83-94.33). relative to 52.46\% (CI: 51.48, 53.44) and 88.61\% (CI: 87.97, 89.23) respectively.

With 250 labels, SCOPE again achieves the highest accuracy with 95.52\%, followed closely by Fix-Match with 94.93\% with overlapping confidence intervals. Both SCOPE and Fix-Match significantly and substantially outperform the other models in the 250 label experiment.
%
%
With 4000 labels, SCOPE achieves 95.82\% which again significantly outperforms all of the other models in this comparison except Fix-Match which achieves a comparable 95.69\%.






Likewise in the CIFAR-100 experients, with 400 labeled samples, SCOPE achieves 57.81\% (CI: 56.83, 58.78) which significantly and substantially outperforms both Mix-Match 32.39\% (CI: 31.47, 33.32) and Fix-Match 50.55\% (CI: 49.57, 51.53).

With 2500 and 10000 labels, SCOPE achieves 72.98\% (CI: 72.10, 73.85) and 78.38\% (CI: 77.56, 79.18) respectively which significantly and substantially outperforms all other models in this comparison except Fix-Match which achieves statistically comparable average accuracy.

Furthermore, the cumulative aggregated confounding error rate for the SCOPE configuration is presented in table \ref{cer-table}, illustrating variations in its values as the nearest neighbor filter utilizes different neighbor counts for establishing pseudolabels of unlabeled samples incorporated into the supervised branch. For CIFAR-10, this experiment demonstrates that with k = 1, the confounding error rate peaks at 2.3\%. However, as the number of neighbors is increased, the confounding error rate stabilizes at 0.97\% when k reaches 6.  Similarly for CIFAR-100, when the k = 1, the confounding error rate reaches its peak of 2.45\% and stabilizes at 0.98\% when k reaches 6.


\begin{table}
	\caption{Confounding Error Rate}
	\label{cer-table}
	\centering
	\begin{tabular}{lll}
		\toprule
		
		K Neighbors & CIFAR-10 & CIFAR-100  \\
		\midrule
		$k=1$ & $2.3\%$ & $2.45\%$ \\
		$k=2$ & $2.13\%$ & $2.21\%$ \\
		$k=3$ & $1.28\%$ & $1.34\%$ \\
		$k=4$ & $1.2\%$ & $1.3\%$  \\
		$k=5$ & $0.98\%$ & $0.99\%$  \\
		$k=6$ & $0.97\%$ & $0.98\%$  \\
            $k=7$ & $0.97\%$ & $0.98\%$  \\
		\bottomrule
	\end{tabular}
\end{table}


		

\begin{table}
	\caption{CIFAR 10 Accuracy Results (Ablation Study)}
	\label{SCOPE-acc-table}
	\centering
	\begin{tabular}{llll}
		\toprule
		Method & 40 labels  & 250 labels  & 4000 labels  \\
		\midrule
		SCOPE with \\ only Gaussian \\ filter & 93.21\% & 95.28\% & 95.69\%   \\
		SCOPE with \\ only K-Nearest \\ Neighbor filter & 93.37\% & 95.39\% & 95.71\%   \\
		\bottomrule
	\end{tabular}
\end{table}

\begin{table}
	\caption{CIFAR 100 Accuracy Results (Ablation Study)}
	\label{SCOPE-acc-table_1}
	\centering
	\begin{tabular}{llll}
		\toprule
		Method & 400 labels  & 2500 labels  & 10000 labels  \\
		\midrule
		SCOPE with \\ only Gaussian \\ filter & 57.30\% & 72.46\% & 78.12\%   \\
		SCOPE with \\ only K-Nearest \\ Neighbor filter & $57.56\%$ & 72.63\% & 78.27\%  \\
		\bottomrule
	\end{tabular}
\end{table}

An additional comparison was performed to observe the performance of SCOPE by using only the Gaussian filter versus SCOPE using only the K-Nearest Neighbor filter.
In CIFAR-10, SCOPE's Gaussian filter attains $93.21\%$ accuracy with 40 labels, $95.28\%$ with 250 labels, and $95.69\%$ with 4000 labels. K-Nearest Neighbor filter yields $93.37\%$, $95.39\%$, and $95.71\%$ accuracy with the same labeled sample counts. Similarly, for CIFAR-100, Gaussian filter achieves $57.30\%$ accuracy with 400 labels, $72.46\%$ with 2500 labels, and $78.12\%$ with 10000 labels. K-Nearest Neighbor filter produces $57.56\%$, $72.63\%$, and $78.27\%$ accuracy. 

\section{Conclusion}

Semi-supervised deep learning is a rapidly evolving field and recent papers have identified the problem of confounding errors as a major limitation of deep pseudolabeling techniques. In this paper, we present a novel meta-learning algorithm, SCOPE, which greatly reduces the prevalence of confounding errors through the incorporation of outlier removal.  Outlier removal helps to address a theoretical deficiency in pseudolabeling relative to EM by integrating knowledge of the probability of unlabeled samples occurring given the model parameters.
%
%
SCOPE significantly improves upon state-of-the-art pseudolabeling results for the semi-supervised CIFAR-10 and CIFAR-100 classification tasks, especially for the challenging task with only 4 labels per category.  Additionally, SCOPE reduces the confounding error rate to below 1\% for both CIFAR-10 and CIFAR-100 datasets when employing $k=6$ contrastive nearest neighbors. These results highlight the potential of outlier suppression in mitigating confounding issues and enhancing the generalization capacity of deep pseudolabeling methods for semi-supervised learning.

\section*{Acknowledgments}

We would like to thank Michael Majurski and Yelena Yesha for their contributions towards this research. This research was funded in part by NSF award \#1747724 IUCRC Center for Advanced Real Time Analytics (CARTA).


{\small
\bibliographystyle{ieee_fullname}
\bibliography{main}
}


\end{document}